\documentclass[10pt, conference, compsocconf, final]{IEEEtran}

\ifCLASSINFOpdf
\else
\fi
\usepackage{graphicx}
\usepackage[export]{adjustbox}
\usepackage[noadjust]{cite}
\usepackage{subfig}
\usepackage{float}

\usepackage{booktabs}
\usepackage{adjustbox}

 \usepackage{algorithmic,algorithm}
 
\usepackage{graphicx,textcomp,xcolor}
\usepackage{multirow, tabularx, ragged2e}
\newcolumntype{C}{>{\Centering\arraybackslash}X}
\usepackage{multicol}
\usepackage{graphicx}

\newcommand\mytab[1]{\begingroup%
    \renewcommand\arraystretch{1}%
    \begin{tabular}[t]{@{}c@{}}#1\end{tabular}%
    \endgroup}

\usepackage{array}
\newcommand{\PreserveBackslash}[1]{\let\temp=\\#1\let\\=\temp}
\newcolumntype{R}[1]{>{\PreserveBackslash\raggedleft}p{#1}}
\newcolumntype{L}[1]{>{\PreserveBackslash\raggedright}p{#1}}
    
\newcolumntype{s}{>{\hsize=.2\hsize}X}

\begin{document}
%
\title{A Computer Vision-assisted Approach to Automated Real-Time Road Infrastructure Management}

\author{\IEEEauthorblockN{Philippe Heitzmann}
\IEEEauthorblockA{
New York City, NY, USA\\
philippe.f.heitzmann@gmail.com}}

\maketitle

\begin{abstract}
Accurate automated detection of road pavement distresses is critical for the timely identification and repair of potentially accident-inducing road hazards such as potholes and other surface-level asphalt cracks. Deployment of such a system would be further advantageous in low-resource environments where lack of government funding for infrastructure maintenance typically entails heightened risks of potentially fatal vehicular road accidents as a result of inadequate and infrequent manual inspection of road systems for road hazards. To remedy this, a recent research initiative organized by the Institute of Electrical and Electronics Engineers ("IEEE") as part of their 2020 Global Road Damage Detection ("GRDC") Challenge published in May 2020 a novel 21,041 annotated image dataset of various road distresses calling upon academic and other researchers to submit innovative deep learning-based solutions to these road hazard detection problems. Making use of this dataset, we propose a supervised object detection approach leveraging You Only Look Once ("YOLO") and the Faster R-CNN frameworks to detect and classify road distresses in real-time via a vehicle dashboard-mounted smartphone camera, producing 0.68 F1-score experimental results ranking in the top 5 of 121 teams that entered this challenge as of December 2021. 
\end{abstract}


\IEEEpeerreviewmaketitle

\section{Introduction}

Reliable road infrastructure plays a critical role in supporting modern economies through the facilitation of the flow of goods and people. As such a 2020 OECD study of 1,266 European regions across 26 EU member countries estimated statistically significant increases of 0.07\%, 0.12\% and 0.17\% in a region's GDP, population and employment figures respectively as a result of a 1\% increase in that same region's road access to GDP, population and labor resources falling within a three hour drive of its borders over a 22-year period \cite{adler2020roads}. This social and economic importance is further highlighted in low- and middle-income countries where World Bank economists estimated in 2017 that a 10\% reduction in road traffic deaths would contribute to a 3.6\% increase in per capita real GDP of these countries over a 24-year period considering the outsized mortality impact of road traffic-related accidents on populations' more economically productive segments \cite{world2017high}. Road networks' developmental importance is further highlighted in the significant share of OECD government budgets earmarked for road maintenance and construction spending annually, with for instance OECD countries devoting in 2016 a combined \$108.5B or 0.19\% of total OECD GDP to road infrastructure maintenance and with the United States alone accounting for an estimated \$53.4B or 49\% of this amount \cite{oecd_roadstats_perc_gdp}. 

\begin{figure}
  \centering
  \includegraphics[width=\linewidth]{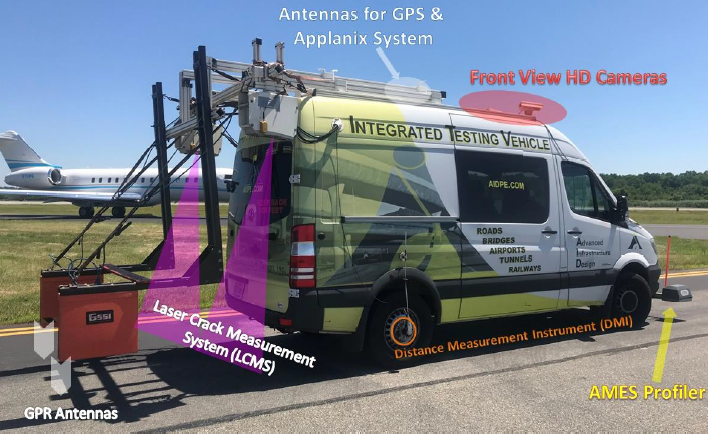}
  \caption{Example of specialized road distress measurement vehicle, developed by \textit{Advanced Infrastructure Design Inc.}, showcased in \cite{njdot_sdi}} 
  \label{fig:itvvec}
\end{figure}

Despite road networks' critical functions as catalysts for economic development, many governments still rely on either relatively inefficient and inaccurate human visual inspections or relatively expensive and difficult to scale laser- and high definition camera-based systems to carry out surface-level quality evaluations of asphalt roadways in order to identify potentially hazardous pavement distresses such as potholes and cracks that may cause road accidents and endanger motorists \cite{arya2021deep}. For instance as the majority of U.S. Department of Transportation (DOT) state agencies employ government workers or third party contractors to provide annual or biennial estimates of state highway pavement deterioration measuring the prevalence and severity of cracking, patching, faulting and joint deterioration per roadmile for different sections of state highways in order to meet federal reporting requirements mandated under the 2012 Moving Ahead for Progress in the 21st Century Act (MAP-21), inspectors will typically complete these estimates through on-the-ground or windshield visual surveys of road distresses in a way that therefore subjects these calculations to a large element of human error \cite{arya2021deep, usfeddot_tamp, njdot_sdi}. Other methods, such as driving specialized vehicles equipped with various sensors such as laser scanners \cite{zhang2018automatic}, ground penetrating radar ("GPR") antennas \cite{nunez2016applications} and high definition cameras \cite{njdot_sdi} \cite{medina2014enhanced} such as shown in Figure \ref{fig:itvvec} along sections of roadway in order to collect high-definition images and 3D reconstructions of the road pavement, are similarly impractical given the significant capital and labor costs required in operating these technologies \cite{arya2021deep}. Given these limitations of existing methods, the relatively more cost-effective alternative of detecting and cataloging road distresses using computer vision algorithms trained on low-cost smartphone-captured pavement images has recently emerged as a subject of interest in academia as public releases of various annotated image datasets of road pavement distresses such as the 2020 GRDC dataset have encouraged further research on the subject.  

\section{Literature Review}

Several deep learning-based methods for the detection of road distresses have been previously proposed with associated annotated image datasets varying in their scale and image subject focus. Zhang et al. \cite{zhang2016road} developed ConvNet- and SVM-based approaches for detecting pavement cracks using a base dataset of 500 pavement crack images of Temple University campus roads augmented 2000x into a final 1M image dataset, producing F-1 scores of 0.90 and 0.74 for each approach respectively. Majidifard et al. \cite{majidifard2020pavement} similarly proposed YOLOv2- and Faster R-CNN-based approaches to road damage detection leveraging a 7,237 annotated image dataset compiled through Google StreetView API queries, yielding 0.84 and 0.65 F-1 scores for each approach respectively. Angulo et al. \cite{angulo2019road} proposed a RetinaNet-based approach to road damage detection making use of a 18,034 annotated image dataset extending Maeda et al.'s \cite{maeda2018road} original 9,053 road distress dataset with additional images collected throughout Italy and Mexico, yielding a model boasting 0.73 F-1 score with 0.5s inference time, allowing for potential real-time road damage detection applications through model deployment on a dashboard-mounted smartphone \cite{angulo2019road}. However as many of these annotated road damage image datasets such as those outlined in \cite{zhang2016road, majidifard2020pavement, angulo2019road} remain private and unavailable to the broader academic research community, the GRDC's public release of this novel image dataset serves an important role in filling this relative dearth of raw image data in this space as further outlined in the following section. 

\section{Road Damage Detection Methodology}

\subsection{2020 GRDC Dataset}

\begin{figure*}
\subfloat[Japan: Longitudinal Crack]{\includegraphics[width = .25\linewidth]{
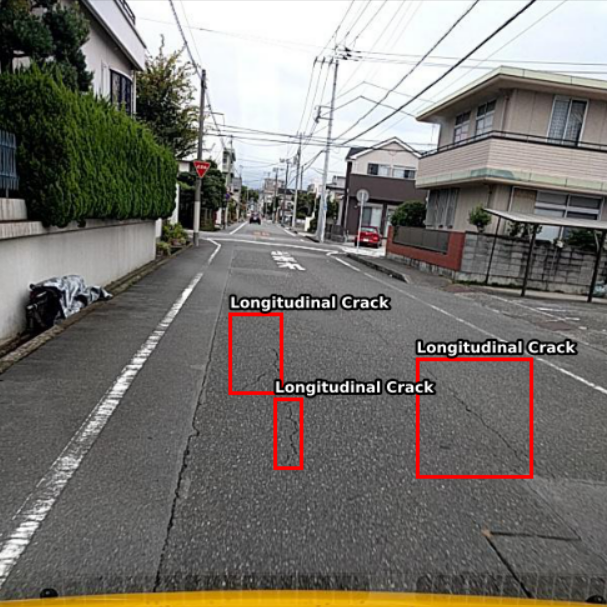}} 
\subfloat[Japan: Lateral Crack]{\includegraphics[width = .25\linewidth]{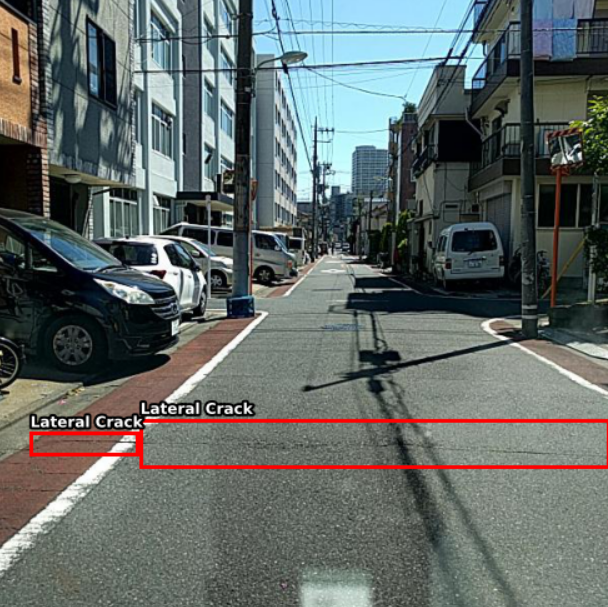}}
\subfloat[Japan: Alligator Crack]{\includegraphics[width = .25\linewidth]{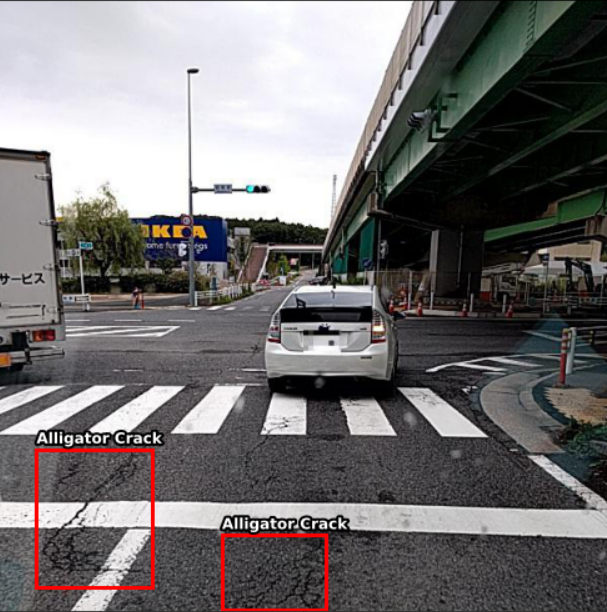}}
\subfloat[Japan: Pothole]{\includegraphics[width = .25\linewidth]{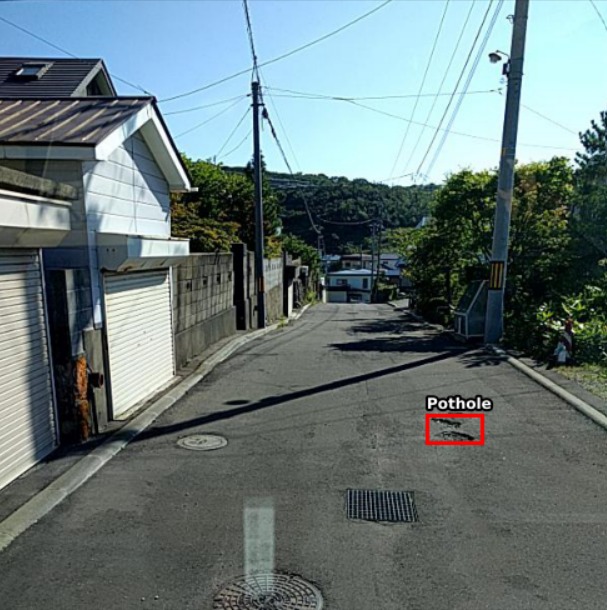}}\\
\subfloat[Czech R.: Longitudinal Crack]{\includegraphics[width = .25\linewidth]{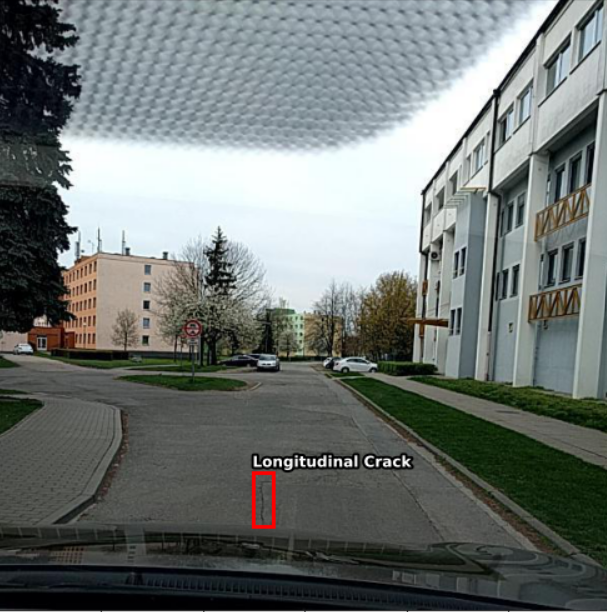}} 
\subfloat[Czech R.: Lateral Crack \& Pothole]{\includegraphics[width = .25\linewidth]{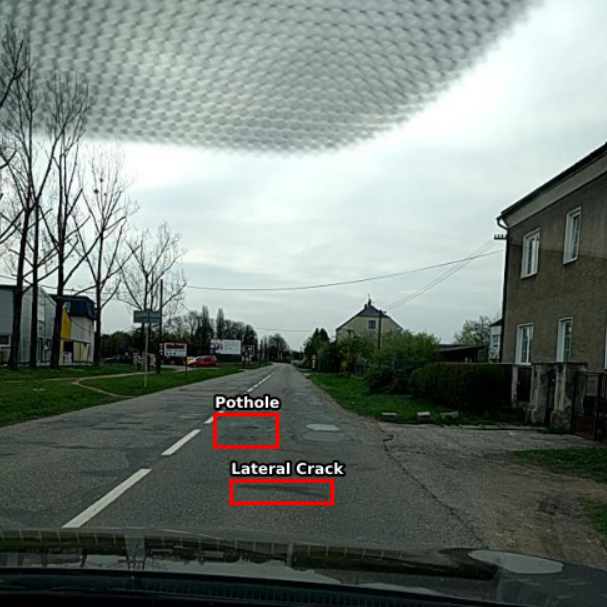}}
\subfloat[India: Alligator Crack]{\includegraphics[width = .25\linewidth]{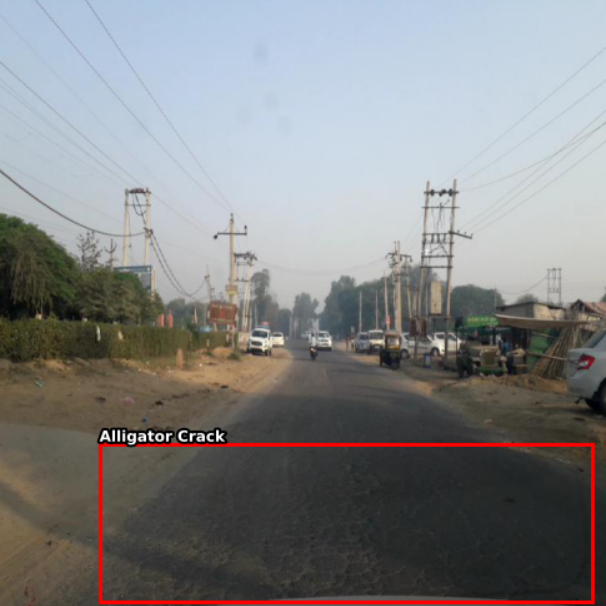}}
\subfloat[India: Pothole]{\includegraphics[width = .25\linewidth]{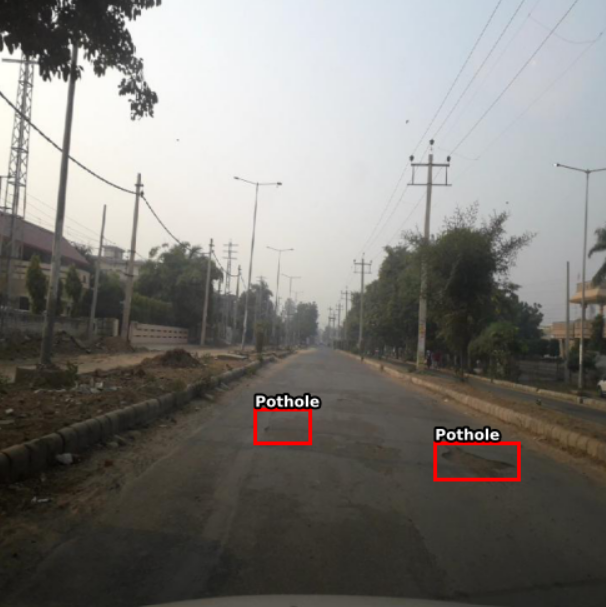}}
\caption{Image Examples of Road Distress Classes}
\label{tab:imdistresses}
\end{figure*}

\begin{figure}
  \centering
  \includegraphics[width=\linewidth]{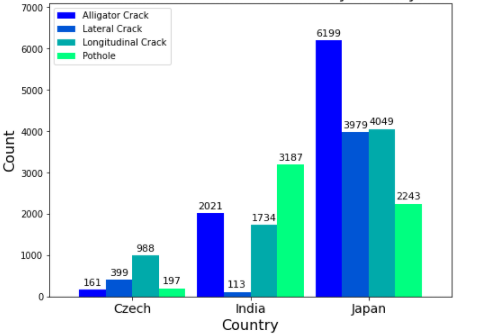}
  \caption{Road Distress Counts by Country} 
  \label{tab:distressfreq}
\end{figure}

The GRDC dataset combines 21,041 road images of pixel sizes 600 x 600 and 720 x 720 captured through a smartphone camera mounted on a vehicle dashboard traveling at an average speed of 25 mph and subsequently hand-annotated by a team of researchers from the Indian Institute of Technology Roorkee and the University of Tokyo ~\cite{arya2021deep}. As the GRDC's stated objective was to develop deep learning models capable of generalizing to predicting road distresses across multiple countries, as opposed to those of a single country as was the focus of the competition's 2018 precursor using 9,053 images collected throughout Japan only, the raw dataset is further divided into 10,506, 7,706 and 2,829 images from Japan, India and the Czech Republic respectively \cite{arya2021deep}. Additionally while the raw dataset provides annotations for a total of eight different road distress classes based on the Japanese Maintenance Guidebooks for Road Pavement \cite{japan_pavement_guidebook}, only the top four classes by frequency counts, namely Longitudinal cracks (class label D00), Lateral Cracks (D10), Alligator Cracks (D20) and Potholes (D40), were considered in the GRDC. Descriptive statistics of the distribution of road distresses and a data dictionary of each are further provided in Table \ref{tab:datadictionary} and Figures \ref{tab:imdistresses} - \ref{tab:distressfreq} respectively. 


\begin{table}[h]
\renewcommand{\arraystretch}{3.5}
\caption{Road Distress Types Overview}
\label{tab:datadictionary}
\centering
\begin{tabular}{|p{1.5cm}|p{1.5cm}|p{4cm}|} 
\cline{1-3}
Distress Name & Distress Code & Sample Image \\
\hline
\hline
Longitudinal Crack & D00 & 
\adjustbox{valign=c}{\includegraphics[width = 4cm]{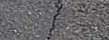}} \\ 
\hline
\hline
Lateral Crack & D10 & 
\adjustbox{valign=c}{\includegraphics[width=4cm]{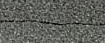}} \\ 
\hline
\hline
Alligator Crack & D20 & 
\adjustbox{valign=c}{\includegraphics[width=4cm]{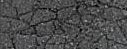}} \\
\hline
\hline
Pothole & D40 & 
\adjustbox{valign=c}{\includegraphics[width=4cm]{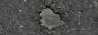}} \\
\hline
\end{tabular}
\end{table}

\subsection{Proposed Approach}

This paper applies both the YOLO ("You Only Look Once") and Faster R-CNN computer vision frameworks to the GRDC dataset in order to explore the relative strengths and weaknesses of applying one- and two-stage detectors to the task at hand respectively. The Faster R-CNN framework developed by Girshick et al. \cite{ren2015faster} may be designated as "two-stage" given its process of first outputting region proposals in an image as candidate regions potentially containing an object of interest before applying a second Region of Interest ("RoI") layer on each region proposal to classify the image and predict bounding box vertices and dimensions within each. While Faster R-CNN boasts higher mAP scores than its predecessor Fast R-CNN and R-CNN models as measured on popular image dataset benchmarks such as MS COCO and PASCAL VOC 2007 and enables real-time detection through approximately 200ms per image test times using GPUs versus 47.0s in the case of R-CNN \cite{ren2015faster} \cite{girshick2015fast}, one-stage detectors such as YOLO were previously developed to overcome some of these foregoing inference time bottlenecks. YOLO may be styled as "one-stage" due to its bypass of Faster R-CNN's region proposal stage in order to divide an input image into a SxS grid of anchor boxes which are then passed through a series of convolutional layers to output a set of \emph{n} bounding boxes with labels. Each of these \emph{n} bounding boxes is then passed through a time-efficient non-maximum suppression ("NMS") algorithm to eliminate bounding boxes with areas of overlap over a certain Intersection over Union ("IoU") threshold \cite{bochkovskiy2020yolov4}. Altogether this one-stage process allows for much improved inference speeds with the latest YOLO implementation of ultralytics-YOLO ("YOLOv5") allowing for per image prediction times in the 7-10ms context using GPUs \cite{yolov5_roboflow}. Given these model architecture and inference time differences we investigated both YOLOv5 in its x (142M trainable parameters) and l (77M parameters) size varieties as well as Faster R-CNN, finding that both YOLOv5-x and l model versions outperformed Faster R-CNN in F1-score and inference time. YOLOv5 was therefore subsequently used as the base model architecture in this approach.

\begin{figure}[h]
  \centering
  \includegraphics[width=\linewidth]{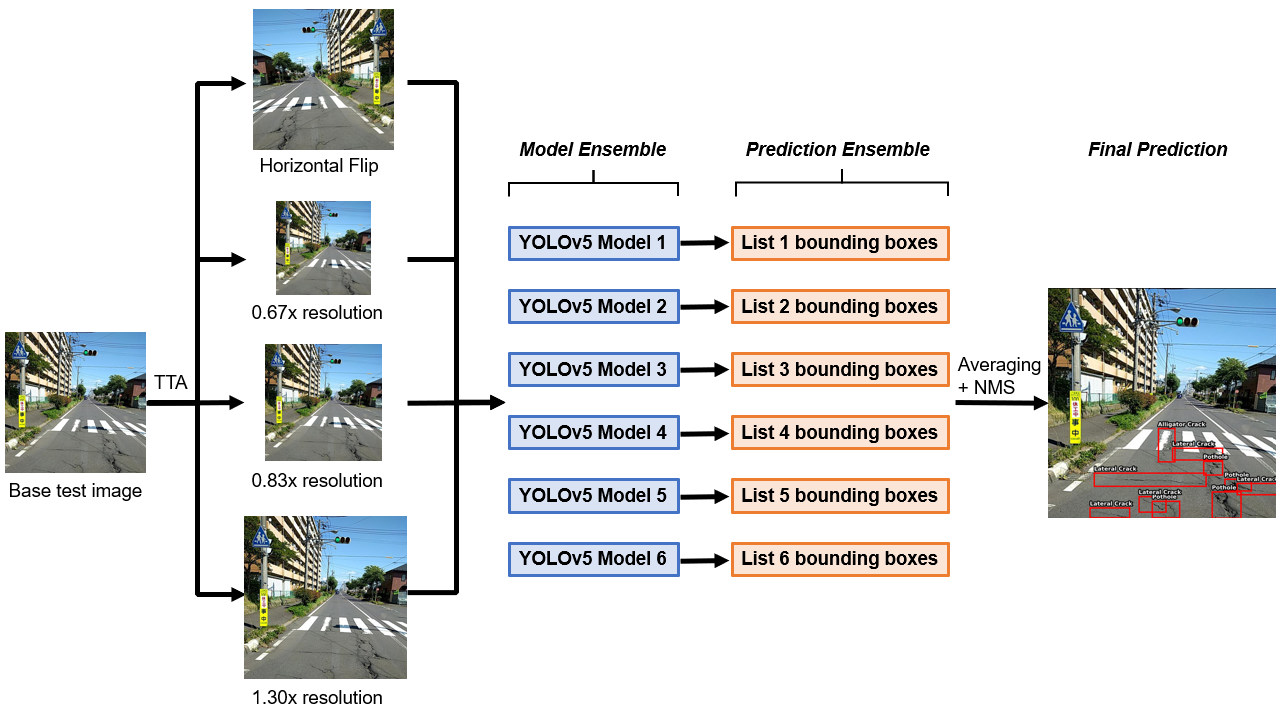}
  \caption{Overview of TTA + EM procedure} 
  \label{fig:tta_em}
\end{figure}

In order to further improve the F1-score performance of this YOLO-based method, the Ensemble Model ("EM") and Test Time Augmentation ("TTA") approaches were further used in the prediction stage. The EM approach \textit{\it ensembles\/} or averages the bounding box predictions of several YOLOv5 models trained with different batch size, learning rate, optimizer and other hyperparameters, with each model's differing kernel patterns learned under its unique set of hyperparameters supplementing those of other included models; as with standard tree-based horizontal or vertical ensembling methods such as Random Forests or Gradient Boosting this has the effect of reducing model prediction variance such that improved accuracy may be achieved \cite{dietterich1995machine}. The tradeoff for this improved accuracy would therefore be increased inference time and reduced model interpretability as no single model would be responsible for the resulting predictions. Similarly this second Test Time Augmentation approach used in this case ensembles individual model predictions on different augmented image versions, derived through horizontal flipping and scaling image resolution 1.30x, 0.83x and 0.67x, of the same base test image. This procedure subsequently filters these five distinct bounding box prediction sets, corresponding to one base and four augmented images, through the NMS procedure based on a selected IoU threshold and a comparison of bounding box confidence scores. This TTA procedure therefore allows for reduced generalization error in its multiple prediction ensembling. Lastly these TTA and EM approaches can be combined such that each k set of base and augmented test images produced through TTA can be fed to each of i EM models in order to yield k * i bounding box prediction sets which are then averaged and filtered through the NMS procedure as detailed in Figure \ref{fig:tta_em}, allowing for increased prediction accuracy through averaging the predictions of several different models across multiple augmented versions of the same base test image.

The GRDC train dataset was further split into 98\% training images and 2\% validation images in order to validate model loss parameter reduction after each training epoch, with these final train and validation sets containing 20,621 and 420 images respectively. In the case of YOLOv5 this base training dataset was further augmented using YOLOv5's standard training augmentation pipeline including horizontal and vertical image flipping and saturation and hue augmentations as detailed in Table \ref{tab:dataaugs}, while this unaltered base dataset was used for Faster R-CNN training. 

\begin{table}[H]
  \centering\footnotesize
\setlength{\tabcolsep}{4pt}
\setlength{\extrarowheight}{3pt}
\caption{YOLOv5 Training Data Augmentations}
\label{tab:dataaugs}
  \begin{tabular}{L{3.5cm}R{3.5cm}} 
    \toprule
    Data Augmentation & Parameter Value  \\
    \midrule
    Hue & 0.7 \\
    Saturation & 0.015 \\
    Translation fraction & 0.1 \\
    Scaling gain & 0.5 \\
    Vertical Flipping & True \\
    Mosaic & True \\
    \bottomrule
  \end{tabular}
\end{table}

\section{Experimental Results}

Per GRDC competition guidelines test scores were derived by submitting through the GRDC's competition website prediction sets for two unreleased test sets containing 2,631 and 2,664 images respectively and sampled following similar country and target class distributions as the training set per the GRDC (datasets "test1" and "test2") \cite{grdc_data_website}. YOLOv5x and l models as well as Faster R-CNN were trained as a first step using standard out-of-the-box training values for learning rate, optimizer, momentum and other hyperparameters, producing F1 scores of 0.52, 0.52 and 0.50 respectively. Additional tuning of batch size and optimizer hyperparameter values showed 8-32 image batch size and stochastic gradient descent with Nesterov accelerated momentum as being optimal in the case of YOLOv5, while SGD with simple momentum and 8-16 batch size were similarly demonstrated as optimal in the Faster R-CNN context. As further tuning YOLOv5x, YOLOv5l and Faster R-CNN models showed superior performance on the part of YOLOv5 as measured by F1 score, the YOLO framework was subsequently adopted as the core of this proposed approach. 

In order to increase model heterogeneity to make this ensemble approach more generalizable, and operating within a maximum inference time constraint of 0.50s in order to theoretically enable real-time detection in the field, several versions of these YOLOv5x and YOLOv5l configured with different batch size and other hyperparameter values were trained and subsequently ensembled. Following this approach an ensemble of six YOLOv5x and YOLOv5l models each trained with 32, 16 and 8 batch sizes for 150 epochs was shown empirically to yield significant improvement over these previous single-model experiments with an F1 score of 0.57, such that this ensemble structure was subsequently selected as the core of this approach. Given it was further observed that per image inference times increased linearly with number of models included in this ensemble this six-model approach producing maximum 0.42ms per image inference times with the vast majority of predictions times falling in the 0.21-0.40ms range was therefore selected to satisfy this self-imposed 0.5s inference time constraint. Following this EM stage, applying TTA augmentations as shown in Figure 3 further allowed for increasing F1 score to 0.59. Finally in order to further improve model prediction performance, an exhaustive grid-search of YOLOv5 NMS and minimum confidence threshold (C) hyperparameter values was conducted in order to ascertain the optimal combination of these hyperparameters, yielding a highest top 5-placing F1 score of 0.68 with C = 0.25 and NMS = 0.999. A summary of all F1 scores produced through this approach for both test1 and test2 datasets is shown below in Tables \ref{tab:test1scores} and  \ref{tab:test2scores}.

\begin{table}[h]
\renewcommand{\arraystretch}{1.75}
\setlength{\tabcolsep}{3pt} 
\scriptsize 
\caption{Test1 F1 Scores of YOLOv5 Model Ensemble varying Confidence Threshold and NMS}
\label{tab:test1scores}
\begin{tabularx}{\columnwidth}{|l|c|*{5}{C|}}
\hline
\multicolumn{2}{|c|}{} & \multicolumn{5}{c|}{Confidence Threshold} \\ 
\cline{3-7} 
\multicolumn{2}{|c|}{} & \textbf{0.1} & \textbf{0.15} & \textbf{0.20} & \textbf{0.25} & \textbf{0.30} \\ 
\hline
\multirow{7}{*}{\mytab{NMS \\ Threshold}}
& \textbf{0.999} & 0.6403 & 0.6645 & 0.6791 & 0.6830 & 0.6786 \\ 
\cline{3-7} 
& \textbf{0.99} & 0.6324 & 0.6591 & 0.6753 & 0.6800 & 0.6763  \\ 
\cline{3-7} 
& \textbf{0.95} & 0.5929 & 0.6248 & 0.6449 & 0.6533 & 0.6537 \\ 
\cline{3-7} 
& \textbf{0.90} & 0.5497 & 0.5862 & 0.6094 & 0.6220 & 0.6260 \\  
\cline{3-7} 
& \textbf{0.85} & 0.5185 & 0.5565 & 0.5826 & 0.5955 & 0.6023 \\ 
\cline{3-7} 
& \textbf{0.80} & 0.4915 & 0.5290 & 0.5572 & 0.5705 & 0.5801  \\
\hline
\end{tabularx}
\end{table}

\begin{table}[h] 
\renewcommand{\arraystretch}{1.75}
\setlength{\tabcolsep}{3pt} 
\scriptsize 
\caption{Test2 F1 Scores of YOLOv5 Model Ensemble varying Confidence Threshold and NMS}
\label{tab:test2scores}
\begin{tabularx}{\columnwidth}{|l|c|*{5}{C|}} 
\hline
\multicolumn{2}{|c|}{} & \multicolumn{5}{c|}{Confidence Threshold} \\ 
\cline{3-7} 
\multicolumn{2}{|c|}{} & \textbf{0.1} & \textbf{0.15} & \textbf{0.20} & \textbf{0.25} & \textbf{0.30} \\ 
\hline
\multirow{7}{*}{\mytab{NMS \\ Threshold}}
& \textbf{0.999} & 0.6306 & 0.6585 & 0.6739 & 0.6770 & 0.6769 \\ 
\cline{3-7} 
& \textbf{0.99} & 0.6240 & 0.6532 & 0.6699 & 0.6744 & 0.6749  \\ 
\cline{3-7} 
& \textbf{0.95} & 0.5828 & 0.6173 & 0.6382 & 0.6477 & 0.6513 \\ 
\cline{3-7} 
& \textbf{0.90} & 0.5416 & 0.5796 & 0.6033 & 0.6155 & 0.6228 \\  
\cline{3-7} 
& \textbf{0.85} & 0.5118 & 0.5508 & 0.5736 & 0.5875 & 0.5970 \\ 
\cline{3-7} 
& \textbf{0.80} & 0.4842 & 0.5211 & 0.5463 & 0.5657 & 0.5757  \\
\hline
\end{tabularx}
\end{table}

\section{System Implementation}

Semi-automated road monitoring systems leveraging computer vision algorithms such as those presented here could be deployed using dashboard-mounted smartphones in order to supplement or potentially replace human visual inspection in either a real-time or offline data processing setting. To further improve recall performance in higher-resource environments, this system could use images taken from several smartphones mounted at different angles in the same vehicle in order to strengthen same-location predictions with different fields of view of the same sections of road. 

\begin{figure}[h]
  \centering
  \includegraphics[width=\linewidth]{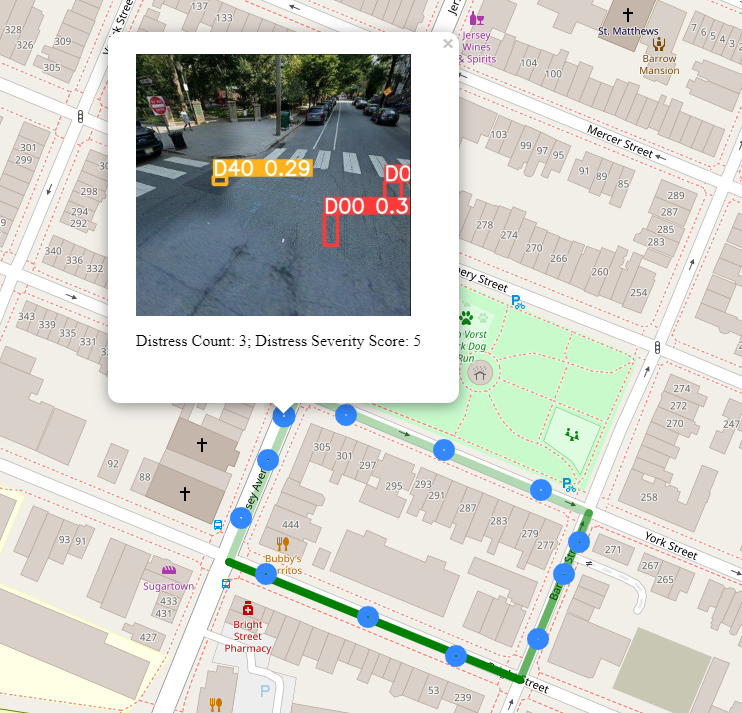}
  \caption{Folium Map of Road Distresses in Paulus Hook neighborhood in Jersey City, NJ} 
  \label{fig:foliummap}
\end{figure}

Furthermore, by using image GPS coordinates automatically embedded in that image file's EXIF data, complete road quality maps of neighborhoods, cities or states could be compiled post-data collection in order to quantify levels of road distress across different road sections, offering a visualization medium to better inform government agencies' road maintenance funding allocation decisions for instance. To demonstrate this, the author created a simple Python folium map of road surface quality in a Paulus Hook neighborhood block in Jersey City, NJ as shown in Figure \ref{fig:foliummap} using road images queried through the Google Street View API and passed to this six-model model ensemble. Leveraging the model's prediction confidence score as a relatively crude proxy for road damage severity, road damage scores can be computed for different sections of road using these road distress frequencies and severities. To facilitate further analysis, this road section-level data could be exported to a tabular format for storing in government agency databases, allowing for comprehensive road analyses across entire cities and states to be performed. 

Other low-cost data collection methods such as smartphone accelerometer data could further reinforce this computer-vision approach such as by providing estimates for road roughness such as proposed in \cite{douangphachanh2013estimation}. As the International Roughness Index (IRI) is another road distress metric commonly monitored by OECD government agencies such as many US Department of Transportation ("DOT") state agencies as part of MAP-21 federal reporting requirements, a computer vision-based model such as that presented in this paper could therefore be supplemented with models regressing IRI on accelerometer data to provide a fuller picture of road quality across both surface quality and roughness \cite{muvcka2016current}. 

\section{Conclusion}

The current relatively elevated costs associated with completing regular and extensive road damage surveys at the local and regional levels through human visual inspection calls for computer vision-assisted monitoring of road infrastructure. This paper put forward a YOLO-based approach to road distress detection using model ensembling and test time augmentation, yielding a 0.68 F1 score on test data placing in the top 5 of 121 teams that entered the 2020 Global Road Detection Challenge as of December 2021. Leveraging this YOLO model ensemble, we furthermore proposed a novel approach to road distress monitoring using several dashboard-mounted smartphones enabling the real-time capture and processing of images and videos of road hazards at different angles. Using a batch of Google Street View API road images with embedded EXIF GPS coordinate data queried for a neighborhood block in Jersey City, NJ, we further demonstrate a simple indexing methodology for quantifying and mapping road surface quality based on distress frequency and severity. As part of future work, we plan to investigate additional methods for improving the cost-effectiveness of road roughness data collection and processing in order to integrate road roughness as an additional dimension to road quality monitoring. 

\section{Acknowledgment}

The author would like to thank Peter Nam for his support in reviewing paper contents.




%

\bibliographystyle{IEEEtran}
\bibliography{main.bib}



\end{document}